\documentclass{article}

\usepackage[cmex10]{amsmath}
\usepackage{amssymb}
\usepackage{gensymb}
\usepackage{textcomp}
\usepackage{pifont}
\usepackage{graphics}
\usepackage{subfigure}

\usepackage{float}
\usepackage{amsmath}
\usepackage{amstext}
\usepackage{amsfonts}
\usepackage[pdftex]{graphicx}
\usepackage{makecell}
\usepackage{array,multirow}
\usepackage{color}
\usepackage{mathtools}
\usepackage{booktabs,tabularx}
\usepackage{spconf,epsfig}
\usepackage{caption}
\usepackage{stfloats}

\newcolumntype{C}[1]{>{\centering\arraybackslash}p{#1}}

\usepackage{enumitem}
\usepackage[linesnumbered,ruled]{algorithm2e}
\usepackage{flushend}
\usepackage{amsthm}
\usepackage{mathrsfs}
\usepackage[table,xcdraw]{xcolor}

\let\OLDthebibliography\thebibliography
\renewcommand\thebibliography[1]{
  \OLDthebibliography{#1}
  \setlength{\parskip}{0pt}
  \setlength{\itemsep}{0pt plus 0.3ex}
}

\begin{document}\sloppy

\title{Multi-modal Document Presentation Attack Detection with Forensics Trace Disentanglement}

\name{Changsheng~Chen$^1$\thanks{This work is sponsored by NSFC 62072313, 62371301, U23B2022, and the CCF-Alibaba Innovative Research Fund for Young Scholars. Z. Yu is the corresponding author.},
        Yongyi~Deng$^1$,
        Liangwei~Lin$^1$,
        Zitong~Yu$^2$*,
        Zhimao~Lai$^3$}
% \author{Changsheng~Chen,
%         Yongyi~Deng,
%         Liangwei~Lin,
%         Zitong~Yu*,
%         Zhimao~Lai
% }% <-this % stops a space
\address{$^1$Guangdong Key Laboratory of Intelligent Information Processing and \\
Shenzhen Key Laboratory of Media Security, Shenzhen University, Shenzhen, China\\
$^2$Great Bay University, Dongguan, China, 
$^3$China People's Police University, Guangzhou, China}
% 

% make the title area
\maketitle

\begin{abstract}
Document Presentation Attack Detection (DPAD) is an important measure in protecting the authenticity of a document image.
However, recent DPAD methods demand additional resources, such as manual effort in collecting additional data or knowing the parameters of acquisition devices.
This work proposes a DPAD method based on multi-modal disentangled traces (MMDT) without the above drawbacks.
We first disentangle the recaptured traces by a self-supervised disentanglement and synthesis network to enhance the generalization capacity in document images with different contents and layouts.
Then, unlike the existing DPAD approaches that rely only on data in the RGB domain, we propose to explicitly employ the disentangled recaptured traces as new modalities in the transformer backbone through adaptive multi-modal adapters to fuse RGB/trace features efficiently.
Visualization of the disentangled traces confirms the effectiveness of the proposed method in different document contents.
Extensive experiments on three benchmark datasets demonstrate the superiority of our MMDT method on representing forensic traces of recapturing distortion.
\end{abstract}
\vspace{-0.2cm}
%\IEEEdisplaynontitleabstractindextext

%\IEEEpeerreviewmaketitle
\vspace{-0.2cm}
\section{Introduction}
\label{sec:Introduction}
\vspace{-0.25cm}
Document images record important information in human society.
With the popularization of digital technology, many public sectors (such as finance and administration) serve society and citizens better through online services.
To ensure the security of personal information and assets, online document (such as ID cards) authentication is required to confirm user identities in many cases.
An Attacker may tamper the document image with photo editing tools and recapture (e.g., print and capture/scan) the tampered version to conceal the forgery traces.
% , recapture, and distribute document images with commodity software and hardware, such as photo editing tools, printing, and imaging devices.
% This poses a risk of misusing document images.
Therefore, Document Presentation Attack Detection (DPAD) is an important task to ensure the authenticity of document images.

Most existing works for recaptured image detection focus on face \cite{liu2021face} and natural images \cite{li2017image}, which are very different from DPAD tasks.
For example, face anti-spoofing detection methods focus on the forensic traces in image depth and material, while no such traces can be found in the task of DPAD.
In recent years, DPAD has attracted a lot of research focus and investigation.
Chen \textit{et al.} \cite{chen2022domain} proposed a three-input Siamese network to extract and compare the features of genuine and recaptured images.
Li \textit{et al.} \cite{li2023two} proposed a two-branch deep neural network that incorporates designed frequency filters and a multi-scale attention fusion module.
%The detection of recaptured images is usually performed only on RGB images.
%It is vulnerable to the unknown domain (e.g., cross-attack types).
%For instance, \cite{chen2022domain} demonstrates good performance under experimental protocols across devices.

Some DPAD methods, such as \cite{chen2022domain}, show limited generalization performance when evaluating document images with different contents and qualities.
To mitigate this issue, \cite{benalcazar2023synthetic, chen2023distortion} proposed various distortion synthesis methods in spatial and spectral domains, respectively, to augment the recaptured features.
The data synthesis method \cite{benalcazar2023synthetic} introduces significant manual effort in collecting real recaptured traces.
The F\&B method \cite{chen2023distortion} requires the knowledge of parameters in the recapturing devices (\textit{i.e.}, resolutions of printing and imaging devices) during the training stage.
Such prior knowledge is unavailable in many datasets, such as DLC2021 \cite{polevoy2022document}.

In this work, we propose the DPAD method with multi-modal disentangled traces (MMDT), which improves the generalization performance of existing DPAD methods \textit{without requiring manual effort in collecting additional data or the knowledge of acquisition devices}.

First, inspired by the spoofing trace disentanglement techniques \cite{liu2020disentangling,liu2022spoof} for face images, we propose to extract the recaptured traces from document images with an end-to-end disentanglement network.
However, disentangling and transferring recaptured traces between document images with different contents and layouts is challenging.
% The forensics traces in recaptured document images of various types are different from those in spoofing faces with fixed facial feature arrangement.
To improve the performance, we define the recaptured traces as blur content and texture components corresponding to the blurring and halftoning distortion, respectively, and design a self-supervised synthesis network that effectively transfers recaptured traces between various documents.

\begin{figure*}
    \centering
    \includegraphics[width=2.0\columnwidth]{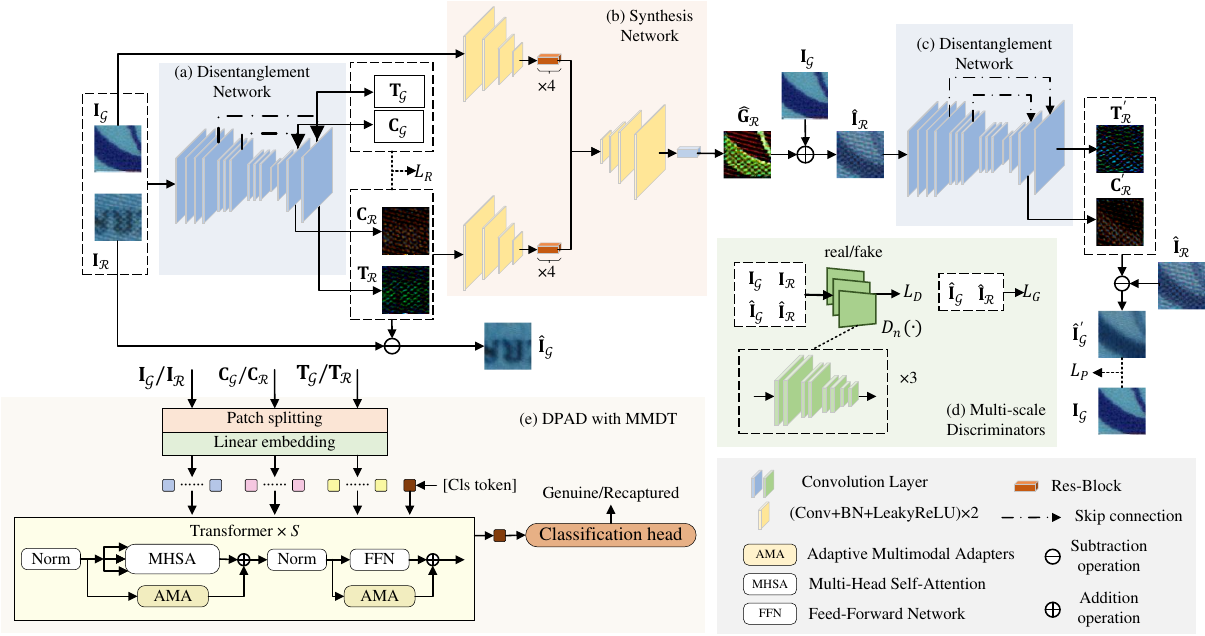}
    \caption{\small The proposed DPAD network with disentangled recaptured traces.
    (a) Our disentanglement network disentangles the blur content $\mathbf{C}$ and the texture $\mathbf{T}$ from an image.
    (b) Our synthesis network synthesizes a recaptured trace $\hat{\mathbf{G}}_{\mathcal{R}}$ by spatially transforming the disentangled $\mathbf{C}_{\mathcal{R}}$ and $\mathbf{T}_{\mathcal{R}}$ in a recaptured image to fit the content of a genuine image $\mathbf{I}_\mathcal{G}$.
    (c) The disentanglement network (same as (a)) disentangles the blur content $\mathbf{C}_R^\prime$ and the texture $\mathbf{T}_R^\prime$ from $\hat{\mathbf{I}}_{\mathcal{R}}$. %$\hat{\mathbf{I}}_{\mathcal{R}}$ is disentangled to obtain the pseudo-genuine image $\hat{\mathbf{I}}^{\prime}_{\mathcal{R}}$.
    % The pseudo-genuine image $\hat{\mathbf{I}}^{\prime}_{\mathcal{R}}$ is obtained by removing recaptured traces from $\hat{\mathbf{I}}_{\mathcal{R}}$.
    (d) The multi-scale discriminators classify real ($\mathbf{I}_{\mathcal{G}};\mathbf{I}_{\mathcal{R}}$) or reconstructed ($\hat{\mathbf{I}}_{\mathcal{G}};\hat{\mathbf{I}}_{\mathcal{R}}$) images.
    (e) The proposed DPAD method with multi-modal disentangled traces (MMDT), which takes $\mathbf{I}_\mathcal{G}$ or $\mathbf{I}_\mathcal{R}$ and its disentangled traces $\mathbf{C},\mathbf{T}$ as the input. 
    During training, we finetuned the AMA and classification head while the parameters of the ViT backbone were frozen.}
    \vspace{-0.5cm}
    \label{fig:Disentanglement}
\end{figure*}

Second, contrary to \cite{benalcazar2023synthetic, chen2023distortion} that encourages recapturing feature extraction implicitly through data augmentation, we propose to explicitly employ the disentangled recaptured traces as new modalities in the DPAD task.
The disentangled traces characterize the blurring and halftoning distortions, which are essential to the DPAD task.
% We employ these traces together with RGB data as multi-modal input to the DPAD model for better generalization.
Thus, our multi-modal DPAD framework inputs the disentangled forensics traces and the RGB data to the transformer backbone and extracts distinctive features from different data modalities by adaptive multi-modal adapters (AMA) \cite{yu2023rethinking}.

To demonstrate the cross-domain performance, we compare our method with a SOTA approach under document images covering various types and different collection devices.
Visualization results confirm that the forensic traces disentangled by the proposed method are consistent with the recapturing distortion in the real document images.
Quantitative results under a cross-domain protocol show that the proposed method outperforms the SOTA approach \cite{chen2023distortion} (using RGB data only) by an improvement of average AUC and EER of 7.97\% and 7.22 percentage points (p.p.), respectively.

The main contributions of this work are as follows.

\noindent $\bullet$ We propose a recaptured trace disentanglement network for document images with various contents by a self-supervised training strategy. Visualizations confirm that the proposed method achieves better performance on document images than an existing approach tailored for spoofing face images.

\noindent $\bullet$ We propose to explicitly fuse RGB features and recaptured traces in our DPAD network via lightweight multi-modal adapters. It achieves better generalization performance than a SOTA approach on the latest DPAD image datasets, without requiring manual effort in collecting additional data or extra knowledge on the datasets.

\noindent $\bullet$ We extend the RSCID dataset \cite{chen2022domain} by incorporating 3,738 low-quality samples with blur and poor lighting, showing more practical application scenarios.

\vspace{-0.2cm}
\section{The Proposed Method}
\label{sec:Proposed Method}
\vspace{-0.25cm}
This section introduces the framework of our forensics trace disentanglement and synthesis network, which is illustrated in Fig. 1.
% We utilize the disentangled traces as novel modalities for recaptured document detection and introduce our DPAD model with multi-modal disentangled traces (MMDT).
To show the contribution of this method, we focus on the trace disentanglement and recaptured image synthesis in Sec.~\ref{subsec:recaptured Trace Disentanglement} and \ref{subsec:Synthesis}.
\vspace{-0.25cm}

\subsection{Recaptured Trace Disentanglement Network}
\label{subsec:recaptured Trace Disentanglement}
\vspace{-0.2cm}
%In this part, we consider the disentanglement process of the recaptured images.
Inspired by the disentanglement of the spoof face \cite{liu2020disentangling,liu2022spoof}, recaptured traces for the DPAD task can be partitioned into multiple components based on their scale, \textit{i.e.}, global traces, blur content traces, and texture traces. 
Global traces refer to color bias and color range disparities induced by recapture.
Blur content traces represent the degradation of texture and characters in the original image caused by recapturing, including blurring distortions in the content, such as text, image, etc. 
Texture traces include the generation of fine textures such as halftone distortions from print-recapture.
% These traces appear across the whole image. 
Different from \cite{liu2020disentangling,liu2022spoof}, we do not consider global traces in our model since color distortion varies significantly across different document images.
Thus, the disentangled recaptured traces $G(\mathbf{I})$ can be defined as
\begin{equation}
    \begin{aligned}
    G(\mathbf{I}) = \lfloor\mathbf{C}\rfloor_N+\mathbf{T},
    \end{aligned}
\end{equation}
\noindent where $\mathbf{C} \in \mathbb{R}^{L \times L \times 3}$ denotes the blur content traces, 
$\mathbf{T} \in \mathbb{R}^{N \times N \times 3}$ is the texture traces ($N>L$, $N$ is the image size), 
and $\lfloor\cdot\rfloor$ is a resizing operation that upsizes $\mathbf{C}$ to the same resolution of $\mathbf{T}$. 

% It should be noted that the difference from \cite{liu2020disentangling} here is our emphasis on $\mathbf{C}$ and $\mathbf{T}$ in recaptured document images.
% Because of the extensive variety in document templates and content, global forensics traces are less stable in recaptured document images.

To supervise the disentanglement network, obtaining the ground truth of recaptured traces from a recaptured document image is not feasible.
Therefore, we set a constraint on the disentangled results to facilitate the training of the disentanglement network.
Specifically, we constraint that the magnitude of disentangled recaptured traces should be close to zero for a genuine image, while that should also be bounded for a recaptured image.
This avoids any unnecessary changes to the document contents.
Thus, the disentanglement process can be formulated as
\begin{equation}
\label{eq:Disentangle}
    G(\mathbf{I}) \approx \underset{\hat{\mathbf{I}}}{\arg \min }\|\mathbf{I}-{\mathbf{\hat{I}}_\mathcal{G}}\|_F, \text { s.t. } \mathbf{I} \in(\mathcal{G} \cup \mathcal{R}),
\end{equation}
\noindent where the domain of genuine and recaptured images are denoted as $\mathcal{G} \subset \mathbb{R}^{N \times N \times 3}$ and $\mathcal{R} \subset \mathbb{R}^{N \times N \times 3}$, respectively, $\mathbf{I}$ is the input image from either domain, $\mathbf{\hat{I}}_\mathcal{G}$ is the reconstructed genuine image, and `$\| \cdot \|_F$' takes the Frobenius norm.

Fig.~\ref{fig:Disentanglement} (a) illustrates the recaptured trace disentanglement network with an encoder-decoder architecture.
The encoder applies a step-wise down-sampling process via convolution layers to the input image (genuine or recaptured), yielding an encoded feature in the latent space $\mathbf{F} \in \mathbb{R}^{28 \times 28 \times 96}$.
The decoder up-samples the feature tensor $\mathbf{F}$ with transposing convolution layers to generate the $\mathbf{T}$ of the same size as the input image.
Skip connections between the encoder and decoder preserve original image information, enhancing information flow and feature reconstruction.
Feature tensors $\mathbf{C} \in \mathbb{R}^{56 \times 56 \times 3}$ are extracted at intermediate decoder layers.
% Here, we employ a regularization loss $\mathit{L}_R$ to restrain the intensity of the disentangled traces.

\vspace{-0.25cm}
\subsection{Self-supervised Synthesis Network}
\label{subsec:Synthesis}
\vspace{-0.2cm}
After obtaining the disentangled recaptured traces extracted from recaptured images, we can apply them to genuine images, thus obtaining reconstructed recaptured samples.
% The pair of genuine images and disentangled traces can then be employed as input for a self-supervised recaptured document image synthesis network. 
However, these recaptured traces contain content-related information associated with the source recaptured image.
% Even when images have the same content template, environmental factors (such as lighting) and human factors (unintentional hand tremors, shooting angles, etc.) often prevent pixel-perfect alignment between the recaptured and genuine images.
Applying the disentangled recaptured traces to genuine images with different content by the existing approaches \cite{liu2022spoof} results in misalignment and strong visual artifacts, as visualized by results in Sec.~\ref{subsec:Visualization}.

% Here, we use a formula to describe the synthesis process. 
%Assuming that $\mathbf{I}_{\mathcal{G}_2}$ is a genuine image, $\hat{\mathbf{I}}_{\mathcal{R}_2}$ is a synthetic recaptured image obtained by recapturing image $\mathbf{I}_{\mathcal{G}_2}$, and $\mathbf{G}_{\mathcal{R}_1}$ ($\mathbf{G}_{\mathcal{R}_1}=(\left\lfloor\mathbf{C}\rfloor_{N}\right)_{\mathcal{R}_{1}}+\mathbf{T}_{\mathcal{R}_1}$) is the traces of another recaptured image.
%The function of the synthesis network can be formulated as:
%\begin{equation}
%    \begin{aligned}
%        \hat{\mathbf{I}}_{\mathcal{R}_{2}} & %\approx\mathbf{I}_{\mathcal{G}_{2}}+\mathbf{G}_{\mathcal{R}_2}\\
%        &=\mathbf{I}_{\mathcal{G}_{2}}+ %S\left[\left(\lfloor\mathbf{C}\rfloor_{N}\right)_{\mathcal{R}_{1}}+\mathbf{T}_{\%mathcal{R}_{1}}, \mathbf{I}_{\mathcal{G}_{2}}\right]
%    \end{aligned}
%\end{equation}
%\noindent where $\mathcal{R}_1,\mathcal{R}_2 \subset \mathcal{R}$ and $ \mathcal{G}_2 \subset \mathcal{G}$ are subsets in the recaptured and genuine image domain, respectively, 
%$\mathcal{R}_{1} \rightarrow \mathcal{R}_{2}$ means the spatial transformation of forensics traces in the recaptured image domain $\mathcal{R}$. 
%$\mathcal{G}_{2}$ and $\mathcal{R}_{2}$ denote the genuine and the recaptured image domains, respectively, with consistent image content information.

To address such limitations, we propose a self-supervised synthesis network to perform spatial transformations on the recaptured traces.
As shown in Fig.~\ref{fig:Disentanglement}~(b), the inputs to the synthesis network include the recaptured features provided by $\mathbf{C}_\mathcal{R}$ and $\mathbf{T}_\mathcal{R}$, along with genuine images providing different content information.
The synthesis network transfers the $\mathbf{C}_\mathcal{R}$ and $\mathbf{T}_\mathcal{R}$ to fit the contents of $\mathbf{I}_{\mathcal{G}}$ and generate the reconstructed recaptured images.
% The encoder in this network employs to process the given input.
Specifically, the genuine image $\mathbf{I}_{\mathcal{G}}$ and recaptured traces $\mathbf{G}_{\mathcal{R}}$ are inputted into the encoder (consisting of convolution layers and batch normalization) respectively derive features $\mathbf{F}_{\mathcal{G}} \in \mathbb{R}^{28 \times 28 \times 256}$ and $\mathbf{F}_{\mathcal{R}} \in \mathbb{R}^{28 \times 28 \times 256}$.
The subsequent Res-Blocks capture and retain features from the inputs to enhance overall network performance.
%The decoder performs the reverse operation of the encoder, and finally, the decoded feature map reduces dimensions through convolution layers to yield the generated spoof trace with different content.
The decoder performs the inverse operation of the encoder.
The decoded feature map undergoes a 3$\times$3 convolutional layer to generate recaptured trace $\mathbf{\hat{G}}_\mathcal{R}$ that has spatially transformed according to the content in the genuine image $\mathbf{I}_{\mathcal{G}}$. 

% with content different from $\mathbf{C}_{\mathcal{R}}$ and $\mathbf{T}_{\mathcal{R}}$.

To facilitate the training of our disentanglement and synthesis networks, we have made an important improvement in the pixel loss $\mathit{L}_P$, which self-supervises the disentanglement process.
It is noted that the disentangled traces from real recaptured image $\mathbf{I}_\mathcal{R}$ and traces from reconstructed recaptured image $\mathbf{\hat{I}}_\mathcal{R}$ contain the same recaptured features but adapted to different image contents.
The pixel loss in existing works \cite{liu2020disentangling, liu2022spoof} cannot directly compare these recaptured traces.
To address this limitation, we approach it from another perspective by computing the pixel loss of real genuine image $\mathbf{I}_{\mathcal{G}}$ and the pseudo-genuine image $\mathbf{\hat{I}}^{\prime}_{\mathcal{G}}$ (output from Fig.~\ref{fig:Disentanglement}~(c)) in a self-supervise fashion.
We define our pixel loss as 
\begin{align}
L_P=\mathbb{E}[\|\mathbf{\hat{I}}_{\mathcal{G}}^{\prime}-\mathbf{I}_\mathcal{G}\|_1],
\end{align}
which verifies the effectiveness of the disentanglement network and enforces the reconstructed recaptured image $\mathbf{{\hat{I}}}_{\mathcal{R}}$ to have the same image content as the real genuine image $\mathbf{I}_{\mathcal{G}}$.
%The supervision of the generated recaptured samples $\hat{\mathbf{I}}_{\mathcal{R}_{2}}$ corresponding to the real genuine samples $\mathbf{I}_{\mathcal{G}_{2}}$ in the generation task is challenging to obtain.

To obtain reconstructed genuine and recaptured images, we design other loss functions following the settings in \cite{liu2020disentangling, liu2022spoof}. 
Specifically, a regularization loss $\mathit{L}_R$ is used to restrain the intensity of the disentangled traces,
a discriminator loss $\mathit{L}_D$ is used for distinguishing between the reconstructed images and the original images.
Moreover, a generator loss $\mathit{L}_G$ is used for ensuring that the recaptured image after recaptured trace disentanglement should be similar to a genuine image. 
% There is a regularization loss $\mathit{L}_R$ to restrain the intensity of the disentangled traces in Fig.~\ref{fig:Disentanglement}~(a).
% We utilize three discriminators with different scales (224,112,56) in Fig.~\ref{fig:Disentanglement}~(d).
% Discriminative supervision $\mathit{L}_D$ is used for distinguishing between the generated images ($\hat{\mathbf{I}}_\mathcal{G};\hat{\mathbf{I}}_\mathcal{R}$) and the original images ($\mathbf{I}_\mathcal{G};\mathbf{I}_\mathcal{R}$).
% Simultaneously, under discriminative supervision, the disentanglement network aims to produce credible images, while the recaptured images after spoof trace disentanglement are classified as non-synthetic images in the loss function $\mathit{L}_G$.

Finally, we define the total loss $\mathit{L}$ of the disentanglement and synthesis network by including the regularizer loss $\mathit{L}_R$, the generator's loss $\mathit{L}_G$, the discriminative loss $\mathit{L}_D$, and the pixel loss $\mathit{L}_P$, which can be formulated as
\begin{equation}
    L = \lambda_1 \cdot L_R + \lambda_2 \cdot L_G + \lambda_3 \cdot L_D + \lambda_4 \cdot L_P,
\end{equation}
\noindent where $\lambda_1,\lambda_2,\lambda_3,\lambda_4$ are the weights to balance the loss from different components.
More details on the loss functions are elaborated in the supplementary.

\vspace{-0.25cm}
\subsection{Multi-modal DPAD Network}
\label{subsec:MultimodalDPAD}
\vspace{-0.2cm}
After disentanglement of the recaptured traces, we propose to incorporate them explicitly in our DPAD model.
Such direct inputs of recaptured traces to the network allow our model to focus on important information for the DPAD task.
This is different from some SOTA approaches that encourage the implicit extraction of recaptured-related features through sample augmentation.

Inspired by the multi-modal methods in improving model performance and robustness \cite{liu2021face,yu2023rethinking}, we apply both the RGB images and the disentangled recaptured traces as multi-modal inputs to our DPAD network.
Specifically, we utilize $\mathbf{C}$ and $\mathbf{T}$ obtained by the disentanglement network (Fig.~\ref{fig:Disentanglement} (a)) as two new modalities of the document image to perform multi-modal DPAD.
This method is called DPAD with multi-modal disentangled trace, abbreviated as MMDT.

% In multi-modal document presentation attack detection, to improve the detection effect, we need each modality to play its role, so the issue of modal fusion is considered.
% In the transformer model, adapters based on fully connected layers primarily concentrate on refining features within individual tokens, overlooking contextual details from nearby tokens and features from different data modes.
% To address these concerns, \cite{yu2023rethinking} expands upon the convolutional adapter (ConvAdapter) to create a multi-modal version for handling document presentation attack detection. Therefore, we add this adaptive multi-modal adapter (AMA) to the transformer.

As illustrated in Fig.~\ref{fig:Disentanglement}~(e), taking the ViT-B16 model as an example, we partition images of these three modalities into patches of $224 \times 224$ pixels to meet the network input size requirements.
Then, the patches of each modality are flattened with a linear projection respectively.
We connect the patch embeddings of each modality and feed the resulting sequence of vectors to the standard transformer encoder in the ViT-B16 model. 
To allow efficient information fusion across different data modalities, we adopt the adaptive multi-modal adapter (AMA) in our transformer backbone \cite{yu2023rethinking}.
AMA is inserted into each transformer encoder as residual connections.
In the whole network, we only finetune the AMA and the classification head while the parameters of other structures are fixed.

%\begin{figure}
    %\centering
    %\includegraphics{img/pic/Multimodal_detection.pdf}
    %\caption{The framework of the ViT finetuning with adaptive multi-modal adapters (AMA).
    %Except that the AMA and classification heads are trainable, other modules use existing pre-trained weights.
    %`MHSA', `FFN', and `GAP' are short for the multi-head self-attention, feed-forward network, and global average pooling, respectively.}
    %\label{fig:Multimodal_detection}
%\end{figure}

\vspace{-0.2cm}
\section{Database and Experimental Results}
\label{sec:Experiments}
\vspace{-0.25cm}
% In this section, we first introduce the recaptured document image datasets employed in our experiment. 
% Next, we present the training details and experimental results of the disentanglement and synthesis network.
% Finally, we present the experimental protocols and quantitative comparison results under each protocol. 
\subsection{Experimental Datasets}
\label{subsec:Dataset}
\vspace{-0.2cm}
In the experiments, we use the following three datasets:

1) \textit{Recaptured Student Card Image Dataset (RSCID)} \cite{chen2022domain}: The document images in this dataset are collected by 11 imaging devices and 3 printing devices.
The templates for these samples are derived from student ID cards of 5 universities.
%All images are divided into two subsets, i.e., dataset $\mathit{D}_1$ and dataset $\mathit{D}_2$, comprising 672 (84 genuine and 588 recaptured) and 432 (48 genuine and 384 recaptured) samples collected by different imaging and printing devices, respectively.
%Due to the choice of acquisition devices, the samples in dataset $\mathit{D}_2$ exhibit higher image quality than those in dataset $\mathit{D}_1$.
Dataset $\mathcal{D}_2$ has 432 (48 genuine and 384 recaptured) samples collected by different imaging and printing devices.
In the experiment, $\mathcal{D}_2$ in RSCID is employed as our training set.

\begin{figure}
    \centering
    \includegraphics[width=0.9\linewidth]{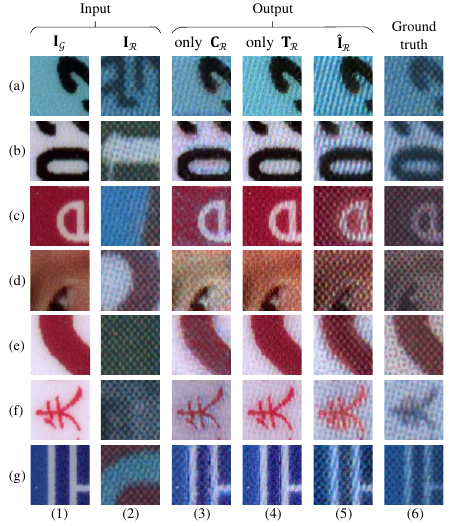}
    \vspace{-0.25cm}
    \caption{\small Examples of reconstructed recaptured images.
    (1) Input genuine images $\mathbf{I}_{\mathcal{G}}$ that provide image content. 
    (2) Input recaptured images $\mathbf{I}_{\mathcal{R}}$ that provide recaptured traces through disentanglement network. 
    (3-5) Recaptured images reconstructed with component $\mathbf{C}_\mathcal{R}$, $\mathbf{T}_{\mathcal{R}}$ and forensics traces $G(\mathbf{I})$, respectively.
    (6) Ground-truth recaptured images with the same contents as $\mathbf{I}_{\mathcal{G}}$ and acquired by the same devices as $\mathbf{I}_{\mathcal{R}}$.}
    \label{fig:Synthesis_result}
    \vspace{-0.5cm}
\end{figure}

2) \textit{Recaptured Document Image Dataset with 162 templates (RDID162)} \cite{chen2023distortion}: This dataset comprises 162 types of documents, including patent certificates, graduation certificates, transcripts, and licenses.
It consists of 162 genuine images and 5,184 recaptured images. 
These image samples (all with resolutions higher than 1500$\times$1500 pixels) are collected through 32 combinations of devices, consisting of 4 printing devices and 8 imaging devices (scanners and cameras).
This dataset is employed as a testing set in our experiment.

3) \textit{RSCID (\textbf{L})}: We extended the original RSCID dataset with low-quality samples (denoted by `\textbf{L}'). 
Specifically, we collected 2,280 genuine and 1,458 recaptured images with low lighting, uneven illumination, bright spots, and laserjet printing distortions. 
This dataset is employed as a testing set in our experiment.
More details of RSCID (\textbf{L}) are included in our supplementary material. 
% The extended dataset is denoted as RSCID (\textbf{L}) in our experiment.
% \begin{figure}
%     \centering
%     \includegraphics[width=0.8\linewidth]{img/pic/Disentanglement_result_comparison_2.pdf}
%     \caption{Examples of forensics trace disentanglement.
%     The (a) is genuine, and the (b)-(g) are recaptured.
%     The (1) column is the input image $\mathbf{I}$ (either genuine or recaptured).
%     The (2) and (3) columns are the forensics traces $G(\mathbf{I})$ and reconstructed images $\hat{I}$ obtained by \cite{liu2020disentangling}, respectively.
%     The (4) and (5) columns are the forensics traces $G(\mathbf{I})$ and reconstructed images $\hat{I}$ obtained by our method, respectively.}
%     \label{fig:Disentanglement_result_comparison}
% \end{figure}

\vspace{-0.25cm}
\subsection{Recaptured Traces Disentanglement and Synthesis}
\label{subsec:Visualization}
\vspace{-0.2cm}
% \subsubsection{Settings in Training Stage}
In this part, we visualize the disentangled recaptured traces and the reconstructed recaptured image produced by our techniques in Sec.~\ref{subsec:recaptured Trace Disentanglement} and \ref{subsec:Synthesis}, respectively. 
This experiment uses the dataset $\mathcal{D}_2$ in RSCID.
The dataset $\mathcal{D}_2$ is divided in an 8:1:1 ratio for the training, validation, and testing sets, respectively.
The proposed method is implemented on TensorFlow with an initial learning rate of $2 \times 10^{-5}$.
Our model is trained in a total of 100K iterations with a batch size of 4.
The parameters $\lambda_1,\lambda_2,\lambda_3,\lambda_4$ are set to be $1,1,1,10$, respectively.
The details on the training process of our network are presented in Sec.~\ref{sec:Supplementary Material} of the supplementary.

% \subsubsection{Visualization of Disentangled Spoofing Traces}
% \label{subsubsec:Visualization}

Columns (2) in Fig.~\ref{fig:Disentanglement_result_comparison} illustrates poor trace disentanglement results on recaptured images by \cite{liu2020disentangling}\footnote{In our experiment, we consider the implementation provided by \cite{liu2020disentangling} since the source code of \cite{liu2022spoof} is not available.}.
In \cite{liu2020disentangling,liu2022spoof}, face shape and the position of facial features are marked with landmarks.
These methods warp the spoofing traces into another facial templates by applying landmark offsets.
However, transferring recaptured traces by warping is not feasible in document images since they have different contents and different landmarks.
Following the same experimental setup, \cite{liu2020disentangling} can be applied to recaptured images but results in poor disentanglement performance.
Due to the limitation of the 3D warping layer, the image disentanglement performance on recaptured images is less effective than that on spoof face images, making it less satisfactory.

As shown in columns (3) of Fig.~\ref{fig:Disentanglement_result_comparison}, we successfully disentangle various recaptured traces.
The disentangled trace $G(\mathbf{I})$ is the difference between the input image and its genuine reconstruction.
As shown in Fig.~\ref{fig:Disentanglement_result_comparison}~(b-g), halftone distortion (especially in the background region) is well detected in our $G(\mathbf{I})$.
Although our method cannot restore image sharpness well when reconstructing genuine images, the model focuses on extracting effective forensic traces.

In Fig.~\ref{fig:Synthesis_result}, we show some examples of reconstructed recaptured images $\mathbf{\hat{I}}_\mathcal{R}$ using the disentangled recaptured traces.
As shown in Fig.~\ref{fig:Synthesis_result}~(a-g), in our synthesis network, these recaptured traces can be transferred onto a genuine image $\mathbf{{I}}_\mathcal{G}$ without any semantic information of the recaptured document $\mathbf{{I}}_\mathcal{R}$.
% and do not carry information about the original image content of the recaptured image.
% The images in columns (2) and (6) are recaptured using the same acquisition device.
Th reconstructed recaptured images $\mathbf{\hat{I}}_\mathcal{R}$ in column (5) generated by our synthesis network exhibit similar forged traces as those real recaptured images in column (6), which are recaptured using the same collection devices as those in column (2).
This also reflects the effectiveness of the traces disentangled by our disentanglement network.
Moreover, trace components $\mathbf{C}_\mathcal{R}$ and $\mathbf{T}_{\mathcal{R}}$ for $G(\mathbf{I})$ can be separately added to a genuine image, creating reconstructed genuine images with different styles.
Compared to existing spoofing trace disentanglement methods for face images \cite{liu2020disentangling, liu2022spoof}, the proposed synthesis network effectively corrects the spatial geometric differences between the source recaptured traces and the target genuine image in the synthesis procedure.

\begin{table*}[htbp]
\begin{minipage}[c]{0.275\linewidth}
\footnotesize
\includegraphics[width=\linewidth]{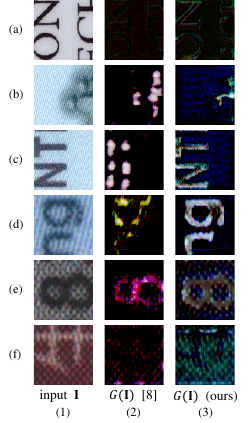}
\vspace{-0.5cm}
\captionof{figure}{\small Comparison of recaptured traces disentangled by \cite{liu2020disentangling} and our disentanglement network with self-supervision. (a) a genuine sample, (b-f) recaptured samples.}
\label{fig:Disentanglement_result_comparison}
\end{minipage}
\hfill
\begin{minipage}[c]{0.7\textwidth}
\centering
\footnotesize
\caption{\small Experiment results in cross-domain protocols. The symbol 
`$\to$' indicates the training set ($\mathcal{D}_2$) on the left and the test set on the right. ALL represents RDID162+RSCID (\textbf{L}). The best performance (with EER as the primary evaluation metric) first and second in each modality are respectively \textbf{bold-faced} and \underline{underlined}.}
\setlength{\tabcolsep}{4.2mm}{
\resizebox{\columnwidth}{!}{%
\begin{tabular}{|ccccccc|}
\hline
\multicolumn{1}{|c|}{\textbf{Protocols}} & \multicolumn{2}{c|}{$\mathcal{D}_2 \to$  RDID162}                                         & \multicolumn{2}{c|}{$\mathcal{D}_2 \to$  RSCID (\textbf{L})}                                         & \multicolumn{2}{c|}{$\mathcal{D}_2 \to$  ALL }                  \\ \hline
\multicolumn{1}{|c|}{\textbf{Methods}}   & \multicolumn{1}{c|}{AUC}             & \multicolumn{1}{c|}{EER}              & \multicolumn{1}{c|}{AUC}             & \multicolumn{1}{c|}{EER}              & \multicolumn{1}{c|}{AUC}             & EER              \\ \hline
\multicolumn{7}{|c|}{\cellcolor[HTML]{E7E6E6}Modality: \textbf{RGB}}                                                                                                                                                                                                      \\ \hline
\multicolumn{1}{|c|}{CDCN \cite{yu2020searching}}               & \multicolumn{1}{c|}{0.6806}          & \multicolumn{1}{c|}{38.78\%}          & \multicolumn{1}{c|}{0.4361}          & \multicolumn{1}{c|}{54.51\%}          & \multicolumn{1}{c|}{0.6803}                & 35.96\%                  \\ \hline
\multicolumn{1}{|c|}{ViT \cite{dosovitskiyimage}}                & \multicolumn{1}{c|}{0.8433}          & \multicolumn{1}{c|}{23.12\%}          & \multicolumn{1}{c|}{0.8590}          & \multicolumn{1}{c|}{23.16\%}          & \multicolumn{1}{c|}{0.8121}                &  26.24\%                \\ \hline
\multicolumn{1}{|c|}{BeiT-B \cite{bao2022beit}}             & \multicolumn{1}{c|}{\textbf{0.8517}} & \multicolumn{1}{c|}{\textbf{22.80\%}} & \multicolumn{1}{c|}{\textbf{0.9339}} & \multicolumn{1}{c|}{\textbf{14.88\%}} & \multicolumn{1}{c|}{\underline{0.8349}} & \underline{24.14\%} \\ \hline
\multicolumn{1}{|c|}{ViT+F\&B \cite{chen2023distortion}}           & \multicolumn{1}{c|}{\underline{0.8457}} & \multicolumn{1}{c|}{\underline{21.65\%}} & \multicolumn{1}{c|}{\underline{0.8900}} & \multicolumn{1}{c|}{\underline{18.30\%}} & \multicolumn{1}{c|}{\textbf{0.8450}} & \textbf{20.92\%} \\ \hline
\multicolumn{1}{|c|}{BeiT-B+F\&B \cite{chen2023distortion}}        & \multicolumn{1}{c|}{0.8043}          & \multicolumn{1}{c|}{28.27\%}          & \multicolumn{1}{c|}{0.8873}          & \multicolumn{1}{c|}{20.49\%}          & \multicolumn{1}{c|}{0.6446}          & 40.32\%          \\ \hline
\multicolumn{7}{|c|}{\cellcolor[HTML]{E7E6E6}Modality: \textbf{RGB+C}}                                                                                                                                                                                                    \\ \hline
\multicolumn{1}{|c|}{MM-CDCN \cite{yu2020multi}}            & \multicolumn{1}{c|}{0.7290}          & \multicolumn{1}{c|}{35.19\%}          & \multicolumn{1}{c|}{0.4921}          & \multicolumn{1}{c|}{43.10\%}          & \multicolumn{1}{c|}{0.7702}          & 29.39\%          \\ \hline
\multicolumn{1}{|c|}{ViT \cite{dosovitskiyimage}}                & \multicolumn{1}{c|}{0.8789}          & \multicolumn{1}{c|}{20.36\%}          & \multicolumn{1}{c|}{0.8635}          & \multicolumn{1}{c|}{21.83\%}          & \multicolumn{1}{c|}{0.8382}          & 25.01\%          \\ \hline
\multicolumn{1}{|c|}{BeiT-B \cite{bao2022beit}}             & \multicolumn{1}{c|}{0.7604}          & \multicolumn{1}{c|}{32.02\%}          & \multicolumn{1}{c|}{0.8298}          & \multicolumn{1}{c|}{24.95\%}          & \multicolumn{1}{c|}{0.8202}          & 26.33\%          \\ \hline
\multicolumn{1}{|c|}{ViT+MMDT}            & \multicolumn{1}{c|}{\textbf{0.8901}} & \multicolumn{1}{c|}{\textbf{19.04\%}} & \multicolumn{1}{c|}{\underline{0.8777}} & \multicolumn{1}{c|}{\underline{20.24\%}} & \multicolumn{1}{c|}{\underline{0.8805}} & \underline{20.51\%} \\ \hline
\multicolumn{1}{|c|}{BeiT-B+MMDT}         & \multicolumn{1}{c|}{\underline{0.8861}} & \multicolumn{1}{c|}{\underline{19.28\%}} & \multicolumn{1}{c|}{\textbf{0.9405}} & \multicolumn{1}{c|}{\textbf{13.16\%}} & \multicolumn{1}{c|}{\textbf{0.8940}} & \textbf{19.04\%} \\ \hline
\multicolumn{7}{|c|}{\cellcolor[HTML]{E7E6E6}Modality: \textbf{RGB+T}}                                                                                                                                                                                                    \\ \hline
\multicolumn{1}{|c|}{MM-CDCN \cite{yu2020multi}}            & \multicolumn{1}{c|}{0.6320}          & \multicolumn{1}{c|}{40.16\%}          & \multicolumn{1}{c|}{0.6240}          & \multicolumn{1}{c|}{42.44\%}          & \multicolumn{1}{c|}{0.8346}          & 24.87\%          \\ \hline
\multicolumn{1}{|c|}{ViT \cite{dosovitskiyimage}}                & \multicolumn{1}{c|}{0.8725}          & \multicolumn{1}{c|}{21.71\%}          & \multicolumn{1}{c|}{0.8850}          & \multicolumn{1}{c|}{18.99\%}          & \multicolumn{1}{c|}{0.8385}          & 24.23\%          \\ \hline
\multicolumn{1}{|c|}{BeiT-B \cite{bao2022beit}}             & \multicolumn{1}{c|}{0.8127}          & \multicolumn{1}{c|}{27.27\%}          & \multicolumn{1}{c|}{0.8638}          & \multicolumn{1}{c|}{21.70\%}          & \multicolumn{1}{c|}{0.8533}          & 22.88\%          \\ \hline
\multicolumn{1}{|c|}{ViT+MMDT}            & \multicolumn{1}{c|}{\textbf{0.9149}} & \multicolumn{1}{c|}{\textbf{15.57\%}} & \multicolumn{1}{c|}{\underline{0.9106}} & \multicolumn{1}{c|}{\underline{16.30\%}} & \multicolumn{1}{c|}{\underline{0.8628}} & \underline{20.84\%} \\ \hline
\multicolumn{1}{|c|}{BeiT-B+MMDT}         & \multicolumn{1}{c|}{\underline{0.8816}} & \multicolumn{1}{c|}{\underline{19.83\%}} & \multicolumn{1}{c|}{\textbf{0.9350}} & \multicolumn{1}{c|}{\textbf{14.98\%}} & \multicolumn{1}{c|}{\textbf{0.9036}} & \textbf{17.53\%} \\ \hline
\multicolumn{7}{|c|}{\cellcolor[HTML]{E7E6E6}Modality: \textbf{RGB+C+T}}                                                                                                                                                                                                  \\ \hline
\multicolumn{1}{|c|}{MM-CDCN \cite{yu2020multi}}            & \multicolumn{1}{c|}{0.7089}          & \multicolumn{1}{c|}{34.53\%}          & \multicolumn{1}{c|}{0.6177}          & \multicolumn{1}{c|}{43.46\%}          & \multicolumn{1}{c|}{0.8147}          & 27.58\%          \\ \hline
\multicolumn{1}{|c|}{ViT \cite{dosovitskiyimage}}                & \multicolumn{1}{c|}{0.8849} & \multicolumn{1}{c|}{19.65\%} & \multicolumn{1}{c|}{0.8999}          & \multicolumn{1}{c|}{18.61\%}          & \multicolumn{1}{c|}{0.8298}          & 23.66\%          \\ \hline
\multicolumn{1}{|c|}{BeiT-B \cite{bao2022beit}}             & \multicolumn{1}{c|}{0.7311}          & \multicolumn{1}{c|}{33.90\%}          & \multicolumn{1}{c|}{0.8356}          & \multicolumn{1}{c|}{25.08\%}          & \multicolumn{1}{c|}{0.8188}                &  26.12\%                \\ \hline
\multicolumn{1}{|c|}{ViT+MMDT}            & \multicolumn{1}{c|}{\textbf{0.9180}} & \multicolumn{1}{c|}{\textbf{16.02\%}} & \multicolumn{1}{c|}{\underline{0.9231}} & \multicolumn{1}{c|}{\underline{16.56\%}} & \multicolumn{1}{c|}{\underline{0.8683}}          & \underline{20.44\%}          \\ \hline
\multicolumn{1}{|c|}{BeiT-B+MMDT}         & \multicolumn{1}{c|}{\underline{0.8740}}          & \multicolumn{1}{c|}{\underline{19.14\%}}          & \multicolumn{1}{c|}{\textbf{0.9452}} & \multicolumn{1}{c|}{\textbf{12.53\%}} & \multicolumn{1}{c|}{\textbf{0.8847}} & \textbf{18.75\%}          \\ \hline
\end{tabular}}
}
\vspace{-0.25cm}
\label{tab:cross_domain_result}
\end{minipage}
\end{table*}

\vspace{-0.25cm}
\subsection{Cross-Domain Experiment}
\label{subsec:Cross-Domain Experiment}
\vspace{-0.2cm}
In this part, we evaluate different DPAD approaches under three experimental protocols, \textit{i.e.}, $\mathcal{D}_2 \to$ RDID162, $\mathcal{D}_2 \to$ RSCID (\textbf{L}) and $\mathcal{D}_2 \to$ ALL (represents RDID162+RSCID (\textbf{L})). 
The training and testing sets cover different acquisition devices and diverse document types.
$\mathcal{D}_2 \to$ RDID162 shows a practical application scenario in which our model is trained on a small and constrained dataset while the testing images in 162 templates are gathered under diverse conditions.
The $\mathcal{D}_2 \to$ RSCID (\textbf{L}) protocol demonstrates the performance under low-quality scenarios.
Dataset $\mathcal{D}_2$ of RSCID is split in an 8:2 ratio to form the training and validation sets, and the RDID162 and RSCID (\textbf{L}) datasets are employed for testing.

% \subsubsection{Implementation Details}
% \label{subsubsec:Implementation Details}
We compare the performances of different approaches with the proposed MMDT in this experiment.
For the single-modal case, we consider one CNN model (CDCN \cite{yu2020searching}) and two transformer backbones (ViT-B16 model \cite{dosovitskiyimage} and BeiT-B model \cite{bao2022beit}). 
For multi-modal cases (RGB+C, RGB+T, RGB+C+T), we use the multi-modal implementations of these models, \textit{i.e.}, multi-modal CDCN \cite{yu2020multi}, multi-modal ViT-B16 \cite{yu2023rethinking} and multi-modal BeiT-B (both with and without MMDT). 
It is noted that \cite{liu2020disentangling} is not considered in this experiment due to the poor performance of recaptured trace disentanglement shown in Sec.~\ref{subsec:Visualization}.

These networks utilize input image patches of size $224 \times 224$ pixels and are trained using the Adam optimizer with a batch size of 64.
ImageNet pre-trained weights are used for our transformer encoder. 
For AMA fine-tuning, the number of original and hidden channels are 768 and 64, respectively. 
Fine-tuning of the model lasts up to 30 epochs with a learning rate of $1 \times 10^{-4}$, a weight decay of 0.05, and cross-entropy loss. 
All methods achieve an image-level decision by a majority voting of the patch-level ones.
Performance is evaluated using the Area Under the ROC Curve (AUC) and Equal Error Rate (EER) metrics.

As shown in Tab.~\ref{tab:cross_domain_result}, for the single-modality approach (RGB-only), the performances across the protocols are not stable.
After performing data augmentation by a SOTA DPAD approach, F\&B \cite{chen2023distortion}, the performance for most backbones yields better results than those without F\&B.
However, under the $\mathcal{D}_2 \to$ ALL protocol, the AUC and EER of the BeiT-B+F\&B model dropped by 0.1903 and 16.18 percentage points (p.p.) compared to those of the BeiT-B model.
%This is because the BeiT model (with 85.85 M parameters) is larger than the ViT-B16 model (with 85.80 parameters), which leads to greater difficulties in the training process only with the RGB modality.
The unstable performance of the BeiT-B model is due to the limitation of the pre-training task in this model \cite{tian2023integrally}.

In the case of multi-modal scenarios (RGB+C, RGB+T, and RGB+C+T), especially for the protocol $\mathcal{D}_2 \to$ ALL, MM-CDCN shows an average performance improvement of 0.1262 and 8.68 p.p. compared to those in RGB-only mode.
This emphasizes that the traces disentangled by our method contribute to enhancing the model's performance in complex scenarios.

For the bimodal scenarios (RGB+C and RGB+T), the performance of the methods is generally improved compared to those in RGB-only mode.
In the settings with RGB+T modality, the average AUC and EER of the methods have increased by 0.0185 and 2.01 p.p., 0.0424 and 3.39 p.p., 0.0933 and 7.42 p.p., respectively, compared to those in RGB-only mode for each protocol ($\mathcal{D}_2 \to$ RDID162, $\mathcal{D}_2 \to$ RSCID (\textbf{L}), $\mathcal{D}_2 \to$ ALL.
Especially for the backbones with MMDT, there is a noticeable performance improvement compared to the SOTA method (\textit{i.e.}, F\&B \cite{chen2023distortion}).

For the tri-modal scenarios, our methods achieve better performance than those in dual-modality mode under $\mathcal{D}_2 \to$ RSCID (\textbf{L}) protocol, while our methods with dual-modal and tri-modal configurations perform comparably under the protocols of $\mathcal{D}_2 \to$ RDID162 and $\mathcal{D}_2 \to$ ALL.
%Especially, for each protocol, the BeiT-B with MMDT improves AUC and EER by 0.0697 and 9.13 p.p., 0.0579 and 7.96 p.p., and 0.2401 and 21.57 p.p., respectively, compared to those of BeiT-B with F\&B.
% The tri-modal setting achieves the best performance under the $\mathcal{D}_2 \to$ RDID162 protocol.
This suggests that, by integrating the recaptured traces $\mathbf{C}$ and $\mathbf{T}$, the trained models are more robust under low-quality scenarios.

\vspace{-0.2cm}
\section{Conclusion}
\label{sec:Conclusion}
\vspace{-0.25cm}
This work proposes a DPAD network, MMDT, via information disentanglement and multi-modal learning techniques.
The experimental results confirm that our MMDT method provides better disentanglement results and achieves better generalization performance under challenging protocols.

In the future, we plan to extend our research toward the disentanglement of different forensic traces, such as forgery.
The forgery traces include artifacts in edge, font, and background textures.
The successful disentanglement of such traces will be helpful in a wide range of document image forensic tasks.

% use section* for acknowledgment
% \section*{Acknowledgment}

% The authors would like to thank...

% Can use something like this to put references on a page
% by themselves when using endfloat and the captionsoff option.
%\ifCLASSOPTIONcaptionsoff
  %\newpage
%\fi

\bibliographystyle{IEEEbib}

\small
\bibliography{reference}

\clearpage
\normalsize
\section{Supplementary Material}
\label{sec:Supplementary Material}
\setcounter{table}{0}
\renewcommand{\thetable}{S\arabic{table}}%
\setcounter{figure}{0}
\renewcommand{\thefigure}{S\arabic{figure}}%

\subsection{The Synthesis Network Details}
\label{subsec:The Network Details}
As shown in Tab.~\ref{tab:Convolution-BatchNorm-ReLU Block} - \ref{tab:Decoder Block}, we introduce the structure of different components in our synthesis network.

%\begin{table}[H]
%\footnotesize
%\centering
%\caption{\small Residual Block (Res-block)}
%\label{tab:res-block}
%\begin{tabular}{|c|c|}
%\hline
%Layer  & input $x$                                                                                                                       \\ \hline
%Conv1  & Conv ($1 \times 1$, 8 channels, activation=LeakyReLU)                                                                                 \\ \hline
%Conv2 & Conv ($3 \times 3$, 8 channels, activation=LeakyReLU)                   \\ \hline
%Conv3  & Conv ($1 \times 1$, 8 channels, no activation)                                                                           \\ \hline
%Addition & \begin{tabular}[c]{@{}c@{}}Element-wise Addition:\\ Add the input x to the output of the final convolution\end{tabular} \\ \hline
%BN  & Batch Normalization: Normalize the output                                                                               \\ \hline
%Activation  & LeakyReLU Activation                   \\ \hline
%\end{tabular}
%\end{table}

\begin{table}[H]
\footnotesize
\centering
\caption{\small Convolution-BatchNorm-LeakyReLU Block (Conv + BN + LeakyReLU)}
\label{tab:Convolution-BatchNorm-ReLU Block}
\vspace{-2.5mm}
\begin{tabular}{|c|c|}
\hline
Layer name    & Layer content                                                                                                     \\ \hline
Conv  & Conv ($3 \times 3$, 32 channels, no activation)                                                       \\ \hline
BN & Batch Normalization: Normalize the output                                                             \\ \hline
Activation & LeakyReLU Activation \\ \hline
\end{tabular}
\end{table}

\begin{table}[H]
\footnotesize
\centering
\caption{\small Encoder Block}
\label{tab:Encoder Block}
\vspace{-2.5mm}
\begin{tabular}{|c|c|}
\hline
Layer name    & Layer content                                                                                                                       \\ \hline
Conv1-1    & Conv-BN-LeakyReLU Block (32 channels)                                                                                  \\ \hline
Conv1-2   & Conv-BN-LeakyReLU Block (32 channels)                                                                                  \\ \hline
Pool1    & \begin{tabular}[c]{@{}c@{}}Conv (3$\times$3, strides=2, activation=LeakyReLU):\\ Increase channels to 64\end{tabular}  \\ \hline
Conv2-1   & Conv-BN-LeakyReLU Block (64 channels)                                                                                  \\ \hline
Conv2-2    & Conv-BN-LeakyReLU Block (64 channels)                                                                                  \\ \hline
Pool2    & \begin{tabular}[c]{@{}c@{}}Conv (3$\times$3, strides=2, activation=LeakyReLU):\\ Increase channels to 128\end{tabular} \\ \hline
Conv3-1  & Conv-BN-LeakyReLU Block (128 channels)                                                                                 \\ \hline
Conv3-2   & Conv-BN-LeakyReLU Block (128 channels)                                                                                 \\ \hline
Pool3    & \begin{tabular}[c]{@{}c@{}}Conv (3$\times$3, strides=2, activation=LeakyReLU):\\ Increase channels to 256\end{tabular} \\ \hline
Conv4-1    & Conv-BN-LeakyReLU Block (256 channels)                                                                                 \\ \hline
Conv4-2 & Conv-BN-LeakyReLU Block (256 channels)                                                                                 \\ \hline
\end{tabular}
\end{table}

\begin{table}[H]
\footnotesize
\centering
\caption{\small Decoder Block}
\label{tab:Decoder Block}
\vspace{-2.5mm}
\begin{tabular}{|c|c|}
\hline
Layer name    & Layer content                                                                                                                                            \\ \hline
Conv1-1    & Conv-BN-LeakyReLU Block (256 channels)                                                                                                       \\ \hline
Conv1-2   & Conv-BN-LeakyReLU Block (256 channels)                                                                                                       \\ \hline
Deconv1    & \begin{tabular}[c]{@{}c@{}}Transpose Conv\\ (3$\times$3, strides=2, activation=LeakyReLU)\end{tabular} \\ \hline
Conv2-1   & Conv-BN-LeakyReLU Block (128 channels)                                                                                                       \\ \hline
Conv2-2    & Conv-BN-LeakyReLU Block (128 channels)                                                                                                       \\ \hline
Deconv2    & \begin{tabular}[c]{@{}c@{}}Transpose Conv\\ (3$\times$3, strides=2, activation=LeakyReLU)\end{tabular} \\ \hline
Conv3-1  & Conv-BN-LeakyReLU Block (64 channels)                                                                                                        \\ \hline
Conv3-2   & Conv-BN-LeakyReLU Block (64 channels)                                                                                                        \\ \hline
Deconv3    & \begin{tabular}[c]{@{}c@{}}Transpose Conv\\ (3$\times$3, strides=2, activation=LeakyReLU)\end{tabular} \\ \hline
Conv4-1    & Conv-BN-LeakyReLU Block (32 channels)                                                                                                        \\ \hline
Conv4-2 & Conv-BN-LeakyReLU Block (32 channels)                                                                                                        \\ \hline
\end{tabular}
\end{table}

\subsection{Discriminators and Loss Function}
\label{subsec:Discriminators and Loss Function}
%The disentanglement and synthesis networks are supervised by multi-scale discriminators in the training process.

\subsubsection{Multi-scale Discriminators}
\label{subsubsec: Multi-scale Discriminators}
As shown in Fig.~\ref{fig:Disentanglement}~(d), we apply discriminative supervision on the reconstructed images to ensure the visual quality of reconstructed images.
Discriminative supervision is used for distinguishing between the reconstructed images ($\hat{\mathbf{I}}_\mathcal{G};\hat{\mathbf{I}}_\mathcal{R}$) and the real images ($\mathbf{I}_\mathcal{G};\mathbf{I}_\mathcal{R}$).
Simultaneously, under discriminative supervision, the disentanglement network aims to produce realistic images, while the recaptured images after recaptured trace disentanglement are classified as real images.
To overcome the limited receptive field of a single discriminator for large-scale images, we utilize multi-scale discriminators $D_n(\cdot), n=\{1, 2, 3\}$ at different resolutions (224 $\times$ 224, 112 $\times$ 112, and 56 $\times$ 56 pixels) to allow the focus on different texture scales.
Each discriminator consists of 8 convolutional layers, including 3 downsampling operations.
% While sharing a similar structure, $D_1$ focuses on fine textures at the highest scale, $D_2$ attends to image content at a medium scale, and $D_3$ operates at the lowest scale.

\subsubsection{The Loss Functions}
\label{subsubsec:loss}
%We introduce the loss functions used in training the whole network.
We introduce the loss functions that are not explicitly provided in the main text.

According to Eq.~\eqref{eq:Disentangle}, we need to minimize the intensity of forensics traces. 
these traces satisfy certain domain conditions. This regularizer loss $\mathit{L}_R$ is written as
\begin{equation}
    L_R=\alpha_1 \cdot \mathbb{E}\left[\left\|G\left(\mathbf{I}_{\mathcal{G}}\right)\right\|_2^2\right]+ \alpha_2 \cdot \mathbb{E} \left[\left\|G\left(\mathbf{I}_{ \mathcal{R}}\right)\right\|_2^2\right],
\end{equation}
\noindent where parameters $\alpha_1$ and $\alpha_2$ are set to be $10$ and $1 \times 10^{-4}$, respectively.
%the `$\to$' symbol denotes the transformation of the image domain.
%$\mathbf{I}_{\mathcal{G} \to \mathcal{G}}$ represents that the domain of real genuine images should remain unchanged.
%$\mathbf{I}_{\mathcal{R} \to \mathcal{G}}$ represents the transformation of real recaptured images into synthesized genuine images through the disentanglement network.

%$\hat{\mathbf{I}}_\mathcal{G},\hat{\mathbf{I}}_\mathcal{R}, \mathbf{I}_\mathcal{G}, \mathbf{I}_\mathcal{R}$ are fed into the discriminators $D_n(\cdot), n=\{1, 2, 3\}$.
Discriminators should distinguish the real images and the reconstructed images. 
For the discriminator supervision, the discriminative loss can be formulated as

\begin{footnotesize} 
    \begin{equation}
        \begin{split}
            L_D=\sum_{n=1,2,3}\left\{\mathbb{E}\left[\left(D_n\left(\mathbf{I}_\mathcal{G}\right)-1\right)^2\right]+ \mathbb{E}\left[\left(D_n\left(\mathbf{\hat{I}}_{\mathcal{G}}\right)\right)^2\right] \right.\\ \left. +\mathbb{E}\left[\left(D_n\left(\mathbf{I}_\mathcal{R}\right)-1\right)^2\right]+\mathbb{E}\left[\left(D_n\left(\mathbf{\hat{I}}_{\mathcal{R}}\right)\right)^2\right]\right\},
            \label{L_D}
        \end{split}
    \end{equation}
\end{footnotesize}

%\noindent where $D_n^{G}(\cdot)$ represents distinguishing between the real and reconstructed genuine samples. $D_n^{R}(\cdot)$ represents distinguishing between the real and reconstructed recaptured samples.
\noindent where the first two terms in the equation represent distinguishing between the real and reconstructed genuine samples, while the last two terms represent distinguishing between the real and reconstructed recaptured samples.
%$\mathit{G}$ and $\mathit{R}$ in the above formula represent the two tasks of the discriminators, respectively. 

The generator's loss function can be expressed as

\begin{footnotesize}
    \begin{equation}
        \begin{split}
            L_G=\sum_{n=1,2,3}\left\{\mathbb{E}\left[\left(D_n\left(\mathbf{\hat{I}}_{\mathcal{G}}\right)-1\right)^2\right] \right.\\ \left.+\mathbb{E}\left[\left(D_n\left(\mathbf{\hat{I}}_{\mathcal{R}}\right)-1\right)^2\right]\right\},
            \label{L_G}
        \end{split}
    \end{equation}
\end{footnotesize}

\noindent where the reconstructed images $\hat{\mathbf{I}}_\mathcal{G},\hat{\mathbf{I}}_\mathcal{R}$ are treated as real images and are discriminated against by the multi-scale discriminators. 

%In the above Eq.~\eqref{L_D} and Eq.~\eqref{L_G}, the supervision of the reconstructed recaptured samples $\hat{I}_{\mathcal{R}}$ corresponding to the real genuine samples in the generation task is challenging to obtain.
%So, we need to give guidance to the synthesis network.
%As shown in Fig.~\ref{fig:Disentanglement} (c), the reconstructed recaptured images must ensure that the pseudo-genuine image $\mathbf{\hat{I}}_{\mathcal{G}}^{\prime}$ is consistent with $\mathbf{I}_\mathcal{G}$.
%Then the loss $\mathit{L}_P$ is formed as:
%\begin{equation}
%    L_P=\mathbb{E}\left[\left\|\mathbf{\hat{I}}_{\mathcal{G}}^{\prime}-\mathbf{I}_\mathcal{G}\right\|_1\right].
%\end{equation}

% The final loss function of the training process is the weighted sum of the above loss functions:

% \begin{equation}
%     L = \lambda_1 L_R+\lambda_2 L_G+\lambda_3 L_P
% \end{equation}

% \noindent where $\lambda_1,\lambda_2,\lambda_3$ are the weights to balance the multitask training.

\begin{figure*}[htbp]
    \centering
    \includegraphics[width=2.0\columnwidth]{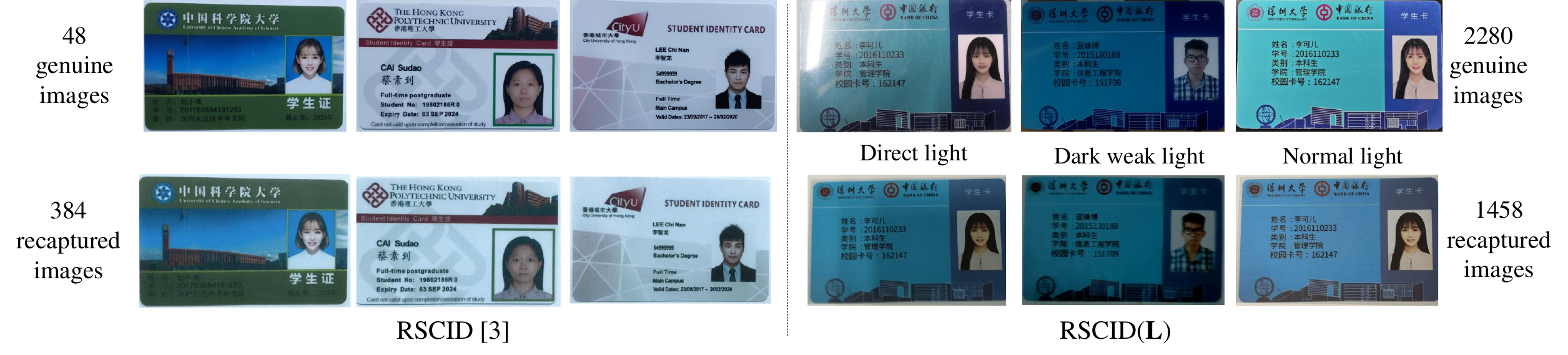}
    \caption{\small Examples of document images in RSCID by \cite{chen2022domain} and our RSCID (\textbf{L}).}
    \vspace{-2.5mm}
    \label{fig:Collected_samples}
\end{figure*}

\subsection{Training of Disentanglement and Synthesis Network}
\label{subsec:Training Part}
The training process of disentanglement and synthesis network can be divided into 3 parts.

\noindent 1) \textbf{Generation part}. 
%The input images (genuine image $\mathbf{I}_\mathcal{G}$ and recaptured image $\mathbf{I}_\mathcal{R}$) are fed to the disentanglement network to disentangle the forensics traces.
%The forensics traces are used to reconstruct the genuine image $\mathbf{\hat{I}}_{\mathcal{G}}$ and synthesize the recaptured image $\mathbf{\hat{I}}_{\mathcal{R}}$.
Input images ($\mathbf{I}_\mathcal{G}$ and $\mathbf{I}_\mathcal{R}$) are fed into the disentanglement and synthesis network with loss functions $L_G$ and $L_R$, resulting in reconstructed images $\mathbf{\hat{I}}_{\mathcal{G}}$ and $\mathbf{\hat{I}}_{\mathcal{R}}$.

\noindent 2) \textbf{Self-supervision part}. 
In this part, the pixel loss $L_P$ is computed between the pseudo-genuine image $\mathbf{\hat{I}}_{\mathcal{G}}^{\prime}$ and the real genuine image $\mathbf{I}_{\mathcal{G}}$, establishing a self-supervised process.
%The frozen disentanglement network is employed to disentangle the reconstructed recaptured image $\mathbf{\hat{I}}_{\mathcal{R}}$, and reconstruct the pseudo-genuine image $\mathbf{\hat{I}}_{\mathcal{G}}^{\prime}$.

%Based on the above two parts, the generator is updated with respect to the regularize loss $\mathit{L}_R$, adversarial loss $\mathit{L}_G$ and pixel loss $\mathit{L}_P$:
%\begin{equation}
%    L_g = \lambda_1 \cdot L_R+\lambda_2 \cdot L_G+\lambda_4 \cdot L_P.
%\end{equation}

\noindent 3) \textbf{Discriminator part}. %$\hat{\mathbf{I}}_\mathcal{G},\hat{\mathbf{I}}_\mathcal{R}, \mathbf{I}_\mathcal{G}, \mathbf{I}_\mathcal{R}$ are fed into the multi-scale discriminators.
The multi-scale discriminators are trained to classify the real ($\mathbf{I}_{\mathcal{G}};\mathbf{I}_{\mathcal{R}}$) or reconstructed ($\hat{\mathbf{I}}_{\mathcal{G}};\hat{\mathbf{I}}_{\mathcal{R}}$) images with the discriminator loss $L_D$.

During the training process, execute parts 1) 2) 3) every even-numbered epoch and parts 1) 2) every odd-numbered epoch until training stops.
It should be noted that, within each epoch, the disentanglement network in Fig.~\ref{fig:Disentanglement}~(a) undergoes parameter updates, while it is frozen in Fig.~\ref{fig:Disentanglement}~(c) after being updated in Fig.~\ref{fig:Disentanglement}~(a).

\begin{table}[t!]
\footnotesize
\centering
\caption{\small The devices used for collecting RSCID (L).}
\label{tab:devices_collect}
\begin{tabular}{|c|c|c|}
\hline
1st Imaging Devices                                                 & Printer Devices                                                                & 2nd Imaging Devices                                                           \\ \hline
\begin{tabular}[c]{@{}c@{}}HUAWEI HONOR 10\\ (24 MP)\end{tabular}   & \multirow{7}{*}{\begin{tabular}[c]{@{}c@{}}Canon C3530\\ 600 DPI\end{tabular}} & \multirow{3}{*}{\begin{tabular}[c]{@{}c@{}}Honor 50SE\\ (12 MP)\end{tabular}} \\ \cline{1-1}
\begin{tabular}[c]{@{}c@{}}HUAWEI HONOR 30\\ (40MP)\end{tabular}    &                                                                                &                                                                               \\ \cline{1-1} \cline{3-3} 
\begin{tabular}[c]{@{}c@{}}HUAWEI P30\\ (40 MP)\end{tabular}        &                                                                                & \begin{tabular}[c]{@{}c@{}}Iqoo Z5\\ (12 MP)\end{tabular}                     \\ \cline{1-1} \cline{3-3} 
\begin{tabular}[c]{@{}c@{}}iPhone 14 Pro Max\\ (48 MP)\end{tabular} &                                                                                & \begin{tabular}[c]{@{}c@{}}OPPO K9x\\ (12 MP)\end{tabular}                    \\ \hline
\end{tabular}
\end{table}

\subsection{Our RSCID (\textbf{L}) Dataset}
\label{subsec:RSCID(L)}
%We collect a database RSCID (\textbf{L}) for our testing purpose.
%This is because additional data samples captured in a broader range of distortion scenarios are needed.
%Datasets $\mathcal{D}_1, \mathcal{D}_2$ of RSCID \cite{chen2022domain} lack samples of these distortion types (e.g. noise distortion).

%Since datasets $\mathcal{D}_1$ and $\mathcal{D}_2$ of RSCID were collected only under normal lighting conditions, we need to collect a new dataset RSCID (\textbf{L}) for robust testing of DPAD approaches under various distortion scenarios.
Given that datasets $\mathcal{D}_1$ and $\mathcal{D}_2$ in RSCID were collected only under normal lighting conditions, we find it necessary to collect a new dataset, denoted as RSCID (\textbf{L}), to do the robust evaluation of DPAD approachres under various distortion scenarios.

Tab.~\ref{tab:devices_collect} reports all the devices used in our extended low-quality samples.
%We employ a printer to print the genuine document, followed by imaging it with smartphones of low imaging quality.
The process of collecting recaptured document images follows the rules outlined in \cite{chen2022domain}, except that we consider some poor lighting conditions.
We shoot the documents under different lighting conditions (direct light, dark weak light, and normal light sources), introducing various noise distortions in the captured document images.
Direct light conditions involve exposing the original documents to direct light sources, resulting in captured document images with noticeable glares and shadows. %noticeable .
%Direct light sources include desk lamps, light sources from fluorescent lights, and direct sunlight. 
Dark weak light conditions aim to capture original documents in a dark environment, preserving only dim light sources that do not directly illuminate the original document.
Document images captured under normal light source conditions typically do not exhibit noticeable glare, and the image information is clear and sharp.

Some examples of document images are shown in Fig.~\ref{fig:Collected_samples}.
The low-quality samples consist of student ID cards with 18 different content templates.
There are six additional templates compared to datasets $\mathcal{D}_1$ and $\mathcal{D}_2$, all of which are real student ID cards.
% Due to privacy concerns, we will not display these six templates here.
We manually crop documents from a white background.
The resolutions of the cropped genuine document images range from 1260$\times$793 pixels to 3382$\times$2151 pixels, while those of the cropped recaptured document images range from 820$\times$690 pixels to 3911$\times$1658 pixels.

%\begin{itemize}
%    \item \textit{Photographing smartphone}: Capture images with the smartphone parallel towards the document plane to avoid geometric distortion and rotation.
%    The captured images are saved in JPEG format with the highest quality coefficient.
%    \item \textit{Environment}: Illuminated by an LED lamp and fluorescent lights in the office.
%    \item \textit{Printer}: Set to color mode with the default printing resolution.
%    \item \textit{Cropping}: We manually crop documents from a white background.
%    Because we avoid geometric distortion and rotation in our data collection process, the cropped region contains a document image with a narrow margin.
%\end{itemize}

% that's all folks
\end{document}